\documentclass[10pt,twocolumn,letterpaper]{article}

\usepackage{cvpr}              

\usepackage{graphicx}
\usepackage{amsmath}
\usepackage{amssymb}
\usepackage{booktabs}
\usepackage{paralist}

\usepackage[pagebackref,breaklinks,colorlinks]{hyperref}

\usepackage[capitalize]{cleveref}
\crefname{section}{Sec.}{Secs.}
\Crefname{section}{Section}{Sections}
\Crefname{table}{Table}{Tables}
\crefname{table}{Tab.}{Tabs.}

\def\cvprPaperID{****} 
\def\confName{CVPR }
\def\confYear{2024}

\begin{document}

\title{CAT: Contrastive Adapter Training for Personalized Image Generation}

\author{$^{*}$Jae Wan Park$^{1,2}$ \qquad $^{*}$Sang Hyun Park$^{1,2}$ \qquad $^{*}$Jun Young Koh$^{1,2}$  \qquad $^{*}$Junha Lee$^{1,2}$ \\ \qquad Min Song$^{1,2}$\\
$^{1}$Yonsei University \qquad $^{2}$Onoma AI \\
}

\maketitle
\def\thefootnote{*}\footnotetext{These authors contributed equally to this work}\def\thefootnote{\arabic{footnote}}

\begin{abstract}
The emergence of various adapters\cite{Alpher13, Alpher14,Alpher17,Alpher18,Alpher19},  including Low-Rank Adaptation (LoRA) applied from the field of natural language processing, has allowed diffusion models \cite{LDM}  to personalize image generation at a low cost. However, due to the various challenges including limited datasets and shortage of regularization and computation resources, adapter training often results in unsatisfactory outcomes, leading to the corruption of the backbone model’s prior knowledge. One of the well known phenomena is the loss of diversity in object generation, especially within the same class which leads to generating almost identical objects with minor variations. This poses challenges in generation capabilities. To solve this issue, we present Contrastive Adapter Training (CAT), a simple yet effective strategy to enhance adapter training through the application of CAT loss. Our approach facilitates the preservation of the base model’s original knowledge when the model initiates adapters. Furthermore, we introduce Knowledge Preservation Score (KPS) to evaluate CAT’s ability to keep the former information. We qualitatively and quantitatively compare CAT's improvement. Finally, we mention the possibility of CAT in the aspects of multi-concept adapter and optimization.

\end{abstract}

\section{Introduction}
\label{sec:intro}
The advent of diffusion models like Stable Diffusion \cite{Alpher01,Alpher02,Alpher03} has advanced the text-to-image generation field, meeting the growing demand for personalized image generation. This demand is driven by the integration of text-to-image technology in various production contexts, including comics, illustrations, and animations. Additionally, accurate personalization has spurred the application of diffusion models in other research areas, such as medical image data synthesis and augmentation \cite{medical1, medical2}, and 3D model generation \cite{3d1, 3d2, 3d3}. Despite these progresses, the complexity of achieving successful personalization poses considerable challenges \cite{IPADAPT, MASACTRL, TFCTIG}, primarily due to stringent data requirements and inherently unstable nature of adapters. 
\begin{figure}
\captionsetup{type=figure, skip=2pt, belowskip=-8pt}
\centering
\includegraphics[width=3.3in]{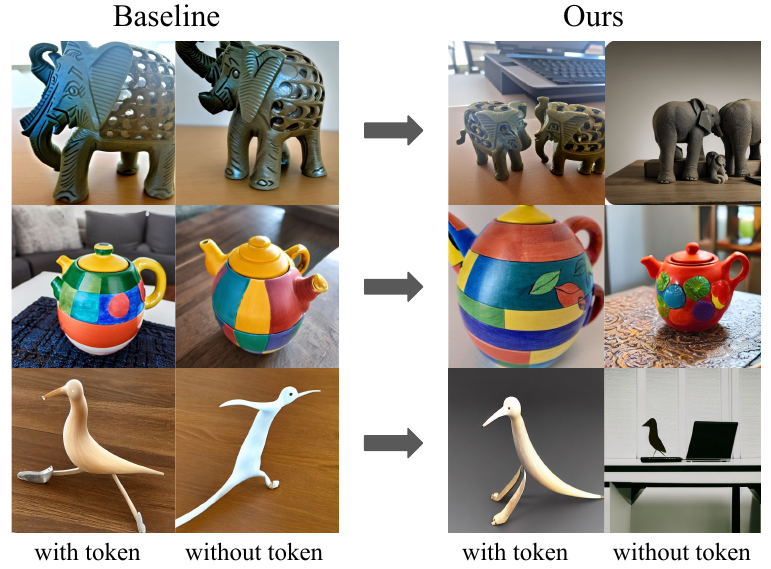}
\caption{
\textbf{Comparison between Baseline \cite{LoRA} and Ours.} We used the following prompts \textit{elephants, a colorful teapot,} and \textit{a thin bird}. Baseline (left) displays knowledge shift and lacking diversity. CAT (right) does not display any knowledge shift and preserves models' ability in diverse manner.
}
\label{fig: Figure 0}
\end{figure}

To lower the burden of dataset requirements and avoid fickle results, various researches utilizing limited datasets have been conducted \cite{Dreambooth, LoRA, TCO, Textual}. Still, the following problems remain unsolved: underfitting and catastrophic forgetting due to overfitting. These problems cause the degradation of generation quality along with unsuccessful identity generation.

We propose a new training pipeline called {\bf C}ontrastive {\bf A}dapter {\bf T}raining that efficiently tackles the aforementioned problems. We adopt a novel form of optimization function that does not require any data augmentation compared to former methods \cite{Dreambooth, TCO}. CAT allows the model to focus contrastively on maintaining the original model's base knowledge by calculating the difference of noise prediction between the original model and adapter without any token conditioning. Lastly, we apply minor modifications to the former metrics and introduce a new metric to measure quantitatively the magnitude of identity generation with knowledge preservation. 

\begin{figure*}
\captionsetup{type=figure, skip=2pt, belowskip=-8pt}
\includegraphics[width=\textwidth]{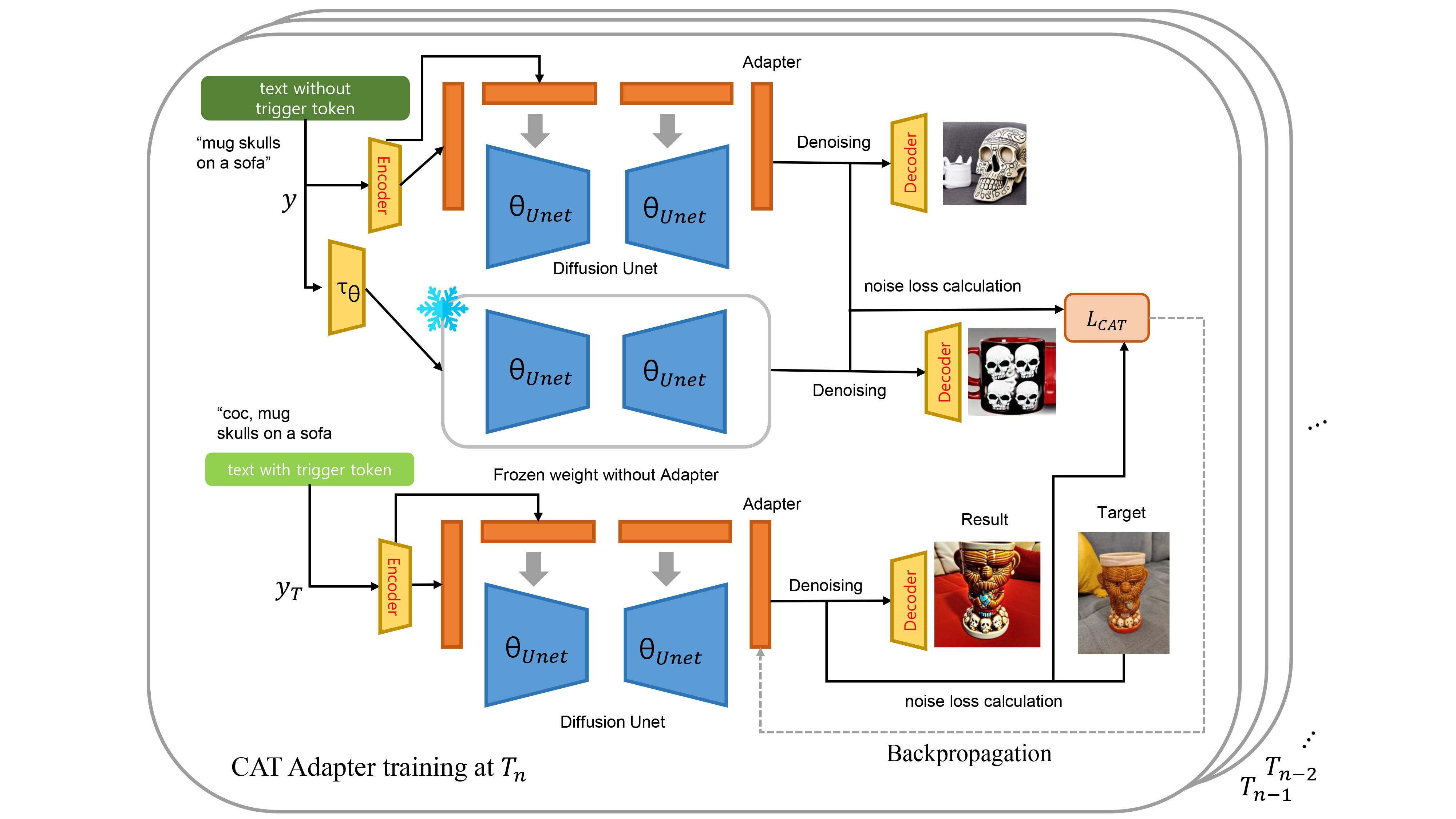}
\caption{
\textbf{The Basic Pipeline of CAT.} The CAT loss between the frozen U-net \cite{Unet} and unconditioned adapter activation is calculated while training the adapter. The Adapters such as LoRA in the attention layers of U-net is depicted as orange boxes. All the figures share the same parameters unless specified.
}
\label{fig: Figure 1}
\end{figure*}

The contributions concerning the proposed approaches are as follows.
\begin{compactitem}
\item We develop a new pipeline that mitigates the underfitting and knowledge corruption problem in consistent generation adaptations.
\item We propose a simplified new metric that assesses the knowledge preservation degree of text-to-image generation adaptations.
\item With the proposed pipeline and metric, we qualitatively and quantitatively show the effectiveness of our proposed methodology.
\end{compactitem}

We also anticipate varied improvements regarding the application and structure of CAT. We mention this part in Section \ref{sec:future}.

\section{Method}
\subsection{Preliminary}


{\bf Latent Diffusion Models} \cite{LDM} are probabilistic architectures that model intractable density functions by sending images to latent spaces. For a certain number, $n$, of steps, $T$, the model passes the latent forward through adding or extracting some noise from Gaussian normal distribution. We also describe this process in the introduction of our method (Fig \ref{fig: Figure 1}). During the train, the model gains the ability to predict how much noise to extract which is the equivalent of modeling the density function. Researches improved the architecture by applying deep learning structures including the implementation of attention architecture \cite{Attention}. This not only has increased the capacity of the model \cite{Alpher02}, but also enabled text-to-image guidance utilizing CLIP embedding \cite{Alpher08}. The mathematical model often denotes the text prompts as $y$, the text encoder as $\tau_{_\theta}$, and the original LDM loss function as
\begin{equation}
    L_{_{LDM}} : =
\mathbb{E}_{\epsilon_{(x)} , y, \epsilon \sim \mathcal{N}(0,1), t} [||\epsilon - \epsilon_{_\theta}(z_{t},t,\tau_{_{\theta}}(y))||^{2}_{2}]
\end{equation}
where $\epsilon$ is the additive Gaussian noise.

\begin{figure*}
\captionsetup{type=figure, skip=0pt, belowskip=0pt}
\begin{tabular}{ccccc}
\textit{with token} &
\textit{with token} &
\textit{with token} &
\textit{with token} &
\textit{with token}\\
\includegraphics[width = 1in]{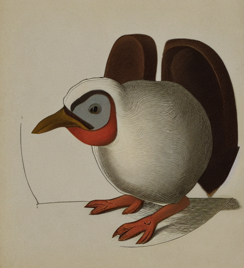} &
\includegraphics[width = 1in]{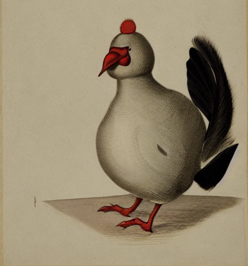} &
\includegraphics[width = 1in]{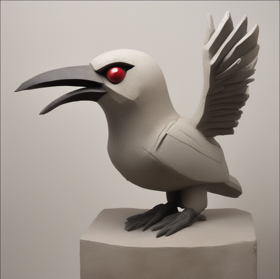} &
\includegraphics[width = 1in]{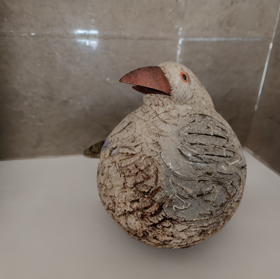} &
\includegraphics[width = 1in]{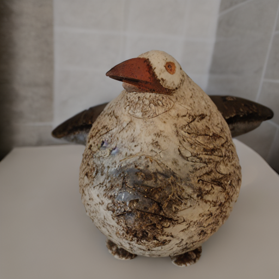}\\
\textit{no token} &
\textit{no token} &
\textit{no token} &
\textit{no token} &
\textit{no token}\\
\includegraphics[width = 1in]{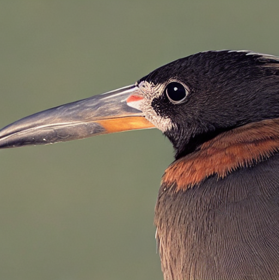} &
\includegraphics[width = 1in]{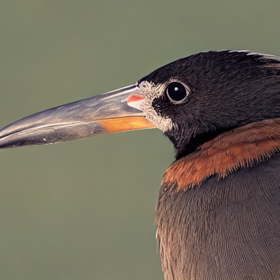} &
\includegraphics[width = 1in]{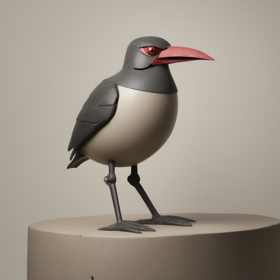} &
\includegraphics[width = 1in]{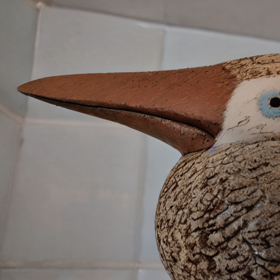} &
\includegraphics[width = 1in]{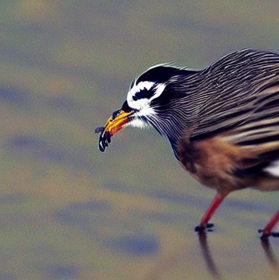}\\
\text{DB} \cite{Dreambooth} &
\text{TI} \cite{Textual} &
\text{TCO} \cite{TCO} &
\text{LoRA} \cite{LoRA} &
\text{CAT(\textbf{Ours})}\\
\end{tabular}
\begin{tabular}{c}
\includegraphics[width = 0.8in]{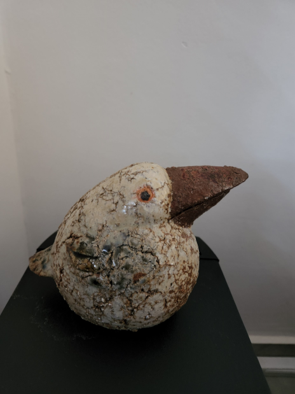} \\
\text{original} 
\end{tabular}
\captionof{figure}{\textbf{Results of text-to-image generation with prompts}: \textit{a round bird staring to the right with its beak closed}. LoRA fails to preserve original knowledge and limits its generation to target image. CAT achieves high fidelity identity generation, while retaining the original knowledge in token-less generation.}
\label{figure 2}
\end{figure*}

{\bf Diffusion Adapters} have rapidly developed in text-to-image generation. Diffusion models have been a significant breakthrough in high quality image generation, yet the realm of generation control is still a field of active research. Many refers to one of the major approaches to enable the control as \textit{personalization}. It is the capability of generating a consistent identity with variety, and to achieve it, adapters are deemed as reasonable solutions.

One of the well received personalization method is Dreambooth \cite{Dreambooth}. The method gains control over consistency using rare-token identifier. It creates a dataset of consistent identity and finetunes the backbone model, mostly diffusion models. The identity images are paired with a rare token: a combination of special, meaningless letters. From this, we denote the initialization of the adapter with $y_{_T}$ which represents prompts with rare tokens added in text guidance. To prevent the mode collapse phenomenon\cite{Alpher12}, or just simply to keep the general capability of the model, it utilizes the self-generated image set of the identity class, called \textit{regularization} set, and constructs a novel preservation loss function.

The idea of linear layer Low-Rank Adaptation \cite{LoRA} has been successfully adapted to text-to-image model's attention layers. It reduced the cost of finetuning in computational aspect, allowing customer-level devices to deal with large language models such as LLAMA and GPT series. \cite{Alpher05, gpt2, gpt3}. Especially in text-to-image models, LoRA has allowed personalization by decomposing mainly U-net \cite{Unet} attention layers during the identity injection steps. LoRA has progressively overwhelmed Dreambooth by reducing the cost of train which solved the underfitting problem, common with Dreambooth training.  

Basic decomposition of attention layers of diffusion model is defined such like

\begin{equation}
    \theta_{n{\times}n}  = \theta_{n{\times}r} \times \theta_{r{\times}n}
    \label{decomposition}
\end{equation}

However, LoRA does not explicitly state or restrict any form of regularization, and small sizes of datasets invoke catastrophic forgetting and cause model's knowledge shift.

Textual Inversion \cite{Textual} took a whole new approach in personalization. The method does not attempt to update the weight of the original model. Rather, it tries to find the embedding in the text embedding space, such as BERT's hot vector space \cite{BERT}. This train targets to generate the identity when the embedding is added to the prompt.

The Chosen One \cite{TCO} utilizes both LoRA and Textual Inversion with K-MEANS++ \cite{k-means} to enhance the targeted identity. It generates a certain number of images like Dreambooth and adopts the machine learning method to cluster identities from the images and augment the intended personalization.

Textual Inversion and The Chosen One introduce unique ways for personalization, but the first still retains the issue of underfitting when the latter tends to overfit the other clustered identity to the intended one when the dataset is limited. Given these complications, we propose a method that sustains the general capacity with better initialization by implementing Contrastive Adapter Training loss.

\subsection{Contrastive Adapter Training}
Following the tradition of Parameter-Efficient Fine-Tuning (PEFT)\cite{PEFT}, all weights are frozen except the target adapter weights. Here, we use LoRA from PEFT as minimal baseline. We adopt Mean Squared Error loss for U-Net's noise prediction between original model and adapted model and accumulate the loss to the original training loss (Fig \ref{fig: Figure 1}). With this approach, CAT minimizes the corruption of the original knowledge and unwanted impact of the adaptation.

Using the rank decomposition notation (\ref{decomposition}) as a part of an adapter initialization, we denote CAT loss function as
\begin{equation}
\begin{gathered}
L_{_{CAT}} : = \\
\begin{aligned}
& \mathbb{E}_{\epsilon_{(x)}, \ y,\ y_{_T}, \ \epsilon \sim \mathcal{N}(0,1),\ t} [||\epsilon - \epsilon_{_\theta}(z_{t},t,\tau_{_{\theta}}(y_{_T}))||^{2}_{2} \\
& + \ \alpha * ||\epsilon_{_\theta}(z_{t},t,\tau_{_\theta}(y)) - \epsilon_{_\theta}(z_{t},t,\tau_{_{\theta_{LoRA}}}(y))||^{2}_{2}]
\end{aligned}
\end{gathered}
\label{cat_eq}
\end{equation}
while the constant $\alpha$ determines the strength of CAT.

\section{Evaluation}
\label{eval}

We have conducted the evaluation regarding CAT mainly by comparing three metrics that uphold our assertion: identity similarity, prompt similarity, and knowledge preservation score. Various studies\cite{Textual, Dreambooth, BASEMCSI} already have accepted the first two, including TCO \cite{TCO}.

We have combined and modified the former metrics to make them more suitable to measure the quality of the proposed method. We named this new metric, \textbf{K}nowledge \textbf{P}reservation \textbf{S}core (KPS). Also, we added some adjustments in them when we aggregate the result to avoid outlier impacts. For notations, see \ref{math}.

{\it Prompt Similarity} evaluates the objective quality of the image. The prompt and generated image without a token are embedded with CLIP encoder, $\theta_{clip}$, and similarity is calculated without normalization. Let the index of the image and caption pair be denoted as $i$, and let $P$ denote the caption part of the pair and $I^{y_{_T}}$ denote the image generated by adapter applied. Then, each pair's score stands for

\begin{equation}
s_{i}^{prompt}= S_{C} (\theta_{clip}(P_{i}),\theta_{clip}(I_{i}^{y_{_T}}))
\label{eq4}
\end{equation}
With this, use harmonic mean to average the score of the whole inference.

\vspace{-15pt}

\begin{equation}
prompt\ score = H(s_{1}^{prompt},s_{2}^{prompt},\dots,s_{i}^{prompt})
\label{prompt}
\end{equation}

{\it Identity Similarity} compares the token generated image from the adapters with the original identity image.  The original image is captioned along with a trigger token to note the identity, and the generated image is guided by this equivalent caption. Two images then are embedded by the CLIP encoder, and cosine similarity is calculated with normalization. Let the score of each pair of images denoted as  

\begin{equation}
s_{i}^{id}= S_{C} (\theta_{clip}(I_{i}^{original}),\theta_{clip}(I_{i}^{y_{_T}}))
\label{eq6}
\end{equation}
Again, for the final score, apply harmonic mean like above.
\begin{equation}
identity \ score = H(s_{1}^{id},s_{2}^{id},\dots,s_{i}^{id})
\label{identity}
\end{equation}

Finally, to evaluate the degree of knowledge preservation, we implement \textit{Knowledge Preservation Score} that compares the images generated by $y$ and $y_{_T}$. Denoted the following,

\vspace{-12pt}

\begin{equation}
s_{i}^{prev}= S_{C} (\theta_{clip}(I_{i}^y),\theta_{clip}(I_{i}^{y_{_T}}))
\label{eq8}
\end{equation}

And since $s_{i}^{prev}$ is normalized within [0,1] and due to the usage of harmonic mean, we can subtract the total aggregation from 1. Adapters that lost the prior knowledge generate similar images whether the prompt has a trigger token or not as shown in Figure \ref{figure 2}. Thus, the score will show that the increase in similarity points out overfitting and knowledge loss from the adapters. 

\begin{multline}
knowledge\ preservation \ score= \\ 1 - H(s_{1}^{prev},s_{2}^{prev},\dots,s_{i}^{prev})
\label{kps}
\end{multline}

We applied this metric to adapters trained in as unbiased environment as possible. We listed the results in Table \ref{tab:Table 1}. For details about the trains and settings, see \ref{train}. 

\vspace{-5pt}

\begin{table}[htbp]
  \centering
  \captionsetup{skip=2pt, belowskip=-11pt} 
  \begin{tabular}{@{}lc@{}lc@{}}
    \toprule
    Method & PS(\ref{prompt}) & \qquad IS(\ref{identity}) & KPS(\ref{kps})\\
    \midrule
    LoRA \cite{LoRA} & 0.2894 & \qquad \bf{0.9056} & 0.0808\\
    Dreambooth \cite{Dreambooth} & 0.2964 & \qquad  0.8394 & 0.0883\\
    Textual Inversion\cite{{Textual}} & \bf{0.3054} & \qquad 0.8405 & 0.0973\\
    TCO \cite{TCO} & 0.2406 & \qquad 0.8416 & 0.0907 \\
    CAT(\bf{Ours}) & 0.2938 & \qquad  0.8968 & \bf{0.1231}\\
    \bottomrule
  \end{tabular}
  \caption{\textbf{Comparison between common adapters}. The prompt score and identity score do not show any drastic gap, however, CAT overwhelms other adaptations in knowledge preservation score.}
  \label{tab:Table 1}
\end{table}

\section{Conclusion}
\label{sec:conclusion}

We accomplish the following two primary objectives. Firstly, we propose Contrastive Adapter Training (Fig \ref{fig: Figure 1} and Eq \ref{cat_eq}), an on-the-fly regularization technique that effectively transforms knowledge shift into knowledge injection. Secondly, we introduce the Knowledge Preservation Score (KPS) as a means to assess the controllability of token generation by utilizing the harmonic mean of similarity scores from trained concepts. Our experiments demonstrate that CAT surpasses competing methods in preserving knowledge, thereby enabling precise control over concept generation. This model notably prioritizes token-specific conditions, avoiding the common pitfall of class-wise mode collapse.

\section{Limitation and Future Works}
\label{sec:future}

We note that we have not included CLIP score-based diversity and fidelity calculation due to its instability \cite{Alpher09, WINO, WVLMBW}. Additionally, we have yet to examine the impact of discrepancies between the model's domain knowledge and the training domain with no thorough investigation on CAT structure and application. In future work, we aim to establish a reliable benchmark for consistent character generation based on our current Knowledge Preservation Score (\ref{kps}). Also, we intend to inspect the impact of CAT (Fig \ref{fig:Figure 4}) and optimizing its structure (Fig \ref{fig:vram_graph}) more carefully. Both tasks will require various examinations such as Figure \ref{fig:graph2}. Mainly, we plan to enhance CAT to support multi-concept training, incorporating per-token loss for individual items demonstrated in Section \ref{MultiConcept}. This expansion holds promise for achieving significant advancements in multi-concept generation \cite{STYLEDROP, OMG, MCCTID} and enhancing the efficiency of personalization strategies.

\section{Acknowledgements}

This work was partly supported by an IITP grant funded by the Korean Government
(MSIT) (No. RS-2020-II201361, Artificial Intelligence Graduate School Program (Yonsei
University)) and was also partly supported by the National Research Foundation of Korea(NRF) grant funded by the Korea government(MSIT) (No. 2022R1A2B5B02002359).

\vspace{10pt}

{\small
\bibliography{egbib}
}

\appendix

\onecolumn

\vspace{-2cm}
\begin{center}
\LARGE\textbf{Appendix}
\label{sec:appendix}
\end{center}
\enlargethispage{\baselineskip}

\begin{figure*}[h]
\captionsetup{type=figure}

\begin{tabular}{ccccc}
\textit{with token} &
\textit{with token} &
\textit{with token} &
\textit{with token} &
\textit{with token}\\
\includegraphics[width = 1in]{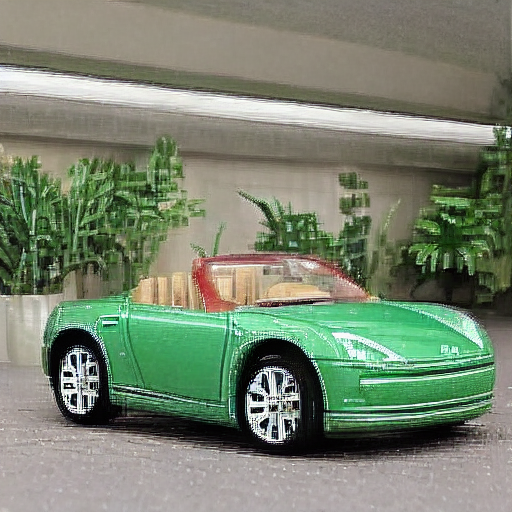} &
\includegraphics[width = 1in]{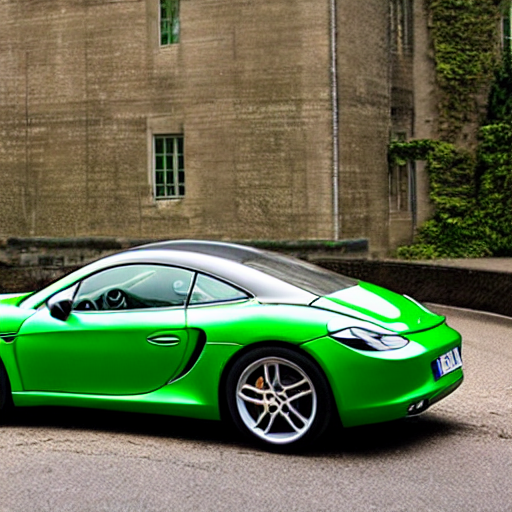} &
\includegraphics[width = 1in]{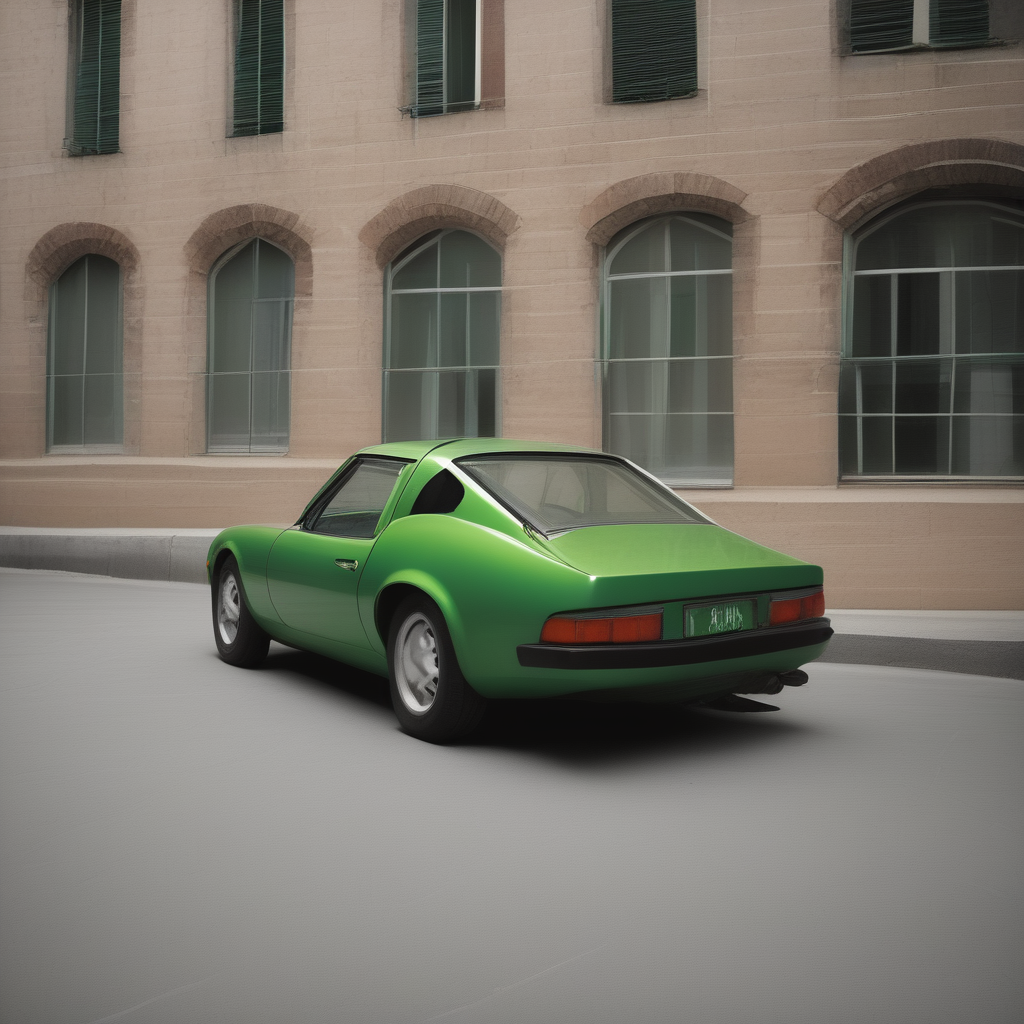} &
\includegraphics[width = 1in]{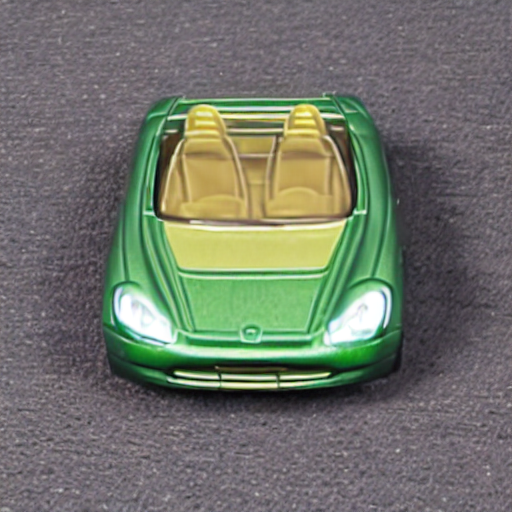} &
\includegraphics[width = 1in]{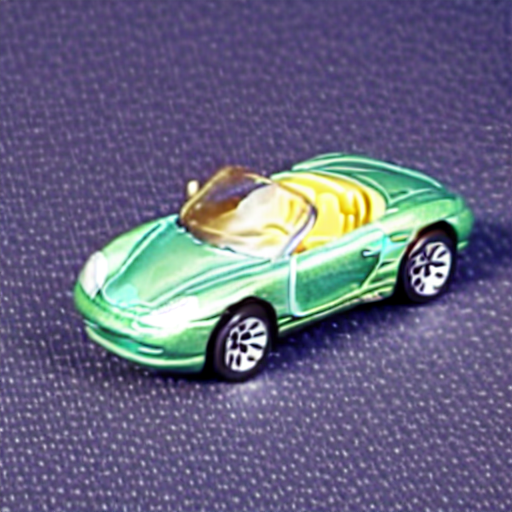}\\
\textit{no token} &
\textit{no token} &
\textit{no token} &
\textit{no token} &
\textit{no token}\\
\includegraphics[width = 1in]{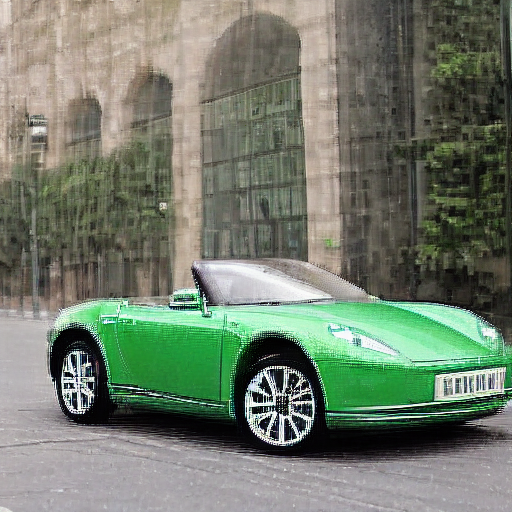} &
\includegraphics[width = 1in]{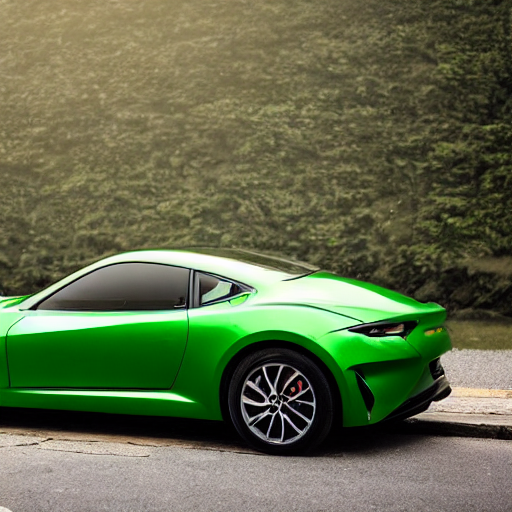} &
\includegraphics[width = 1in]{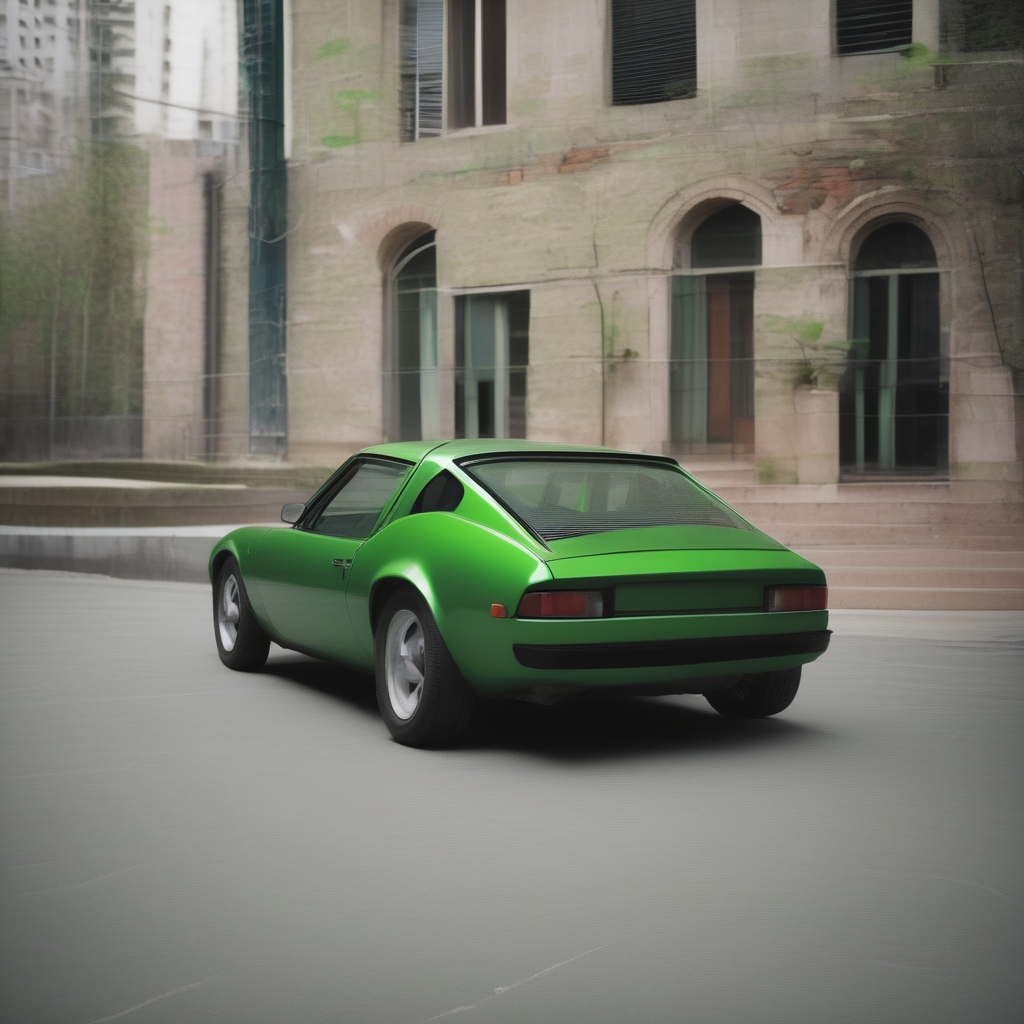} &
\includegraphics[width = 1in]{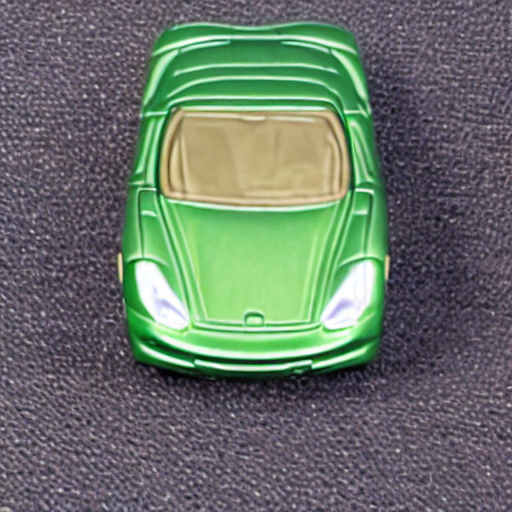} &
\includegraphics[width = 1in]{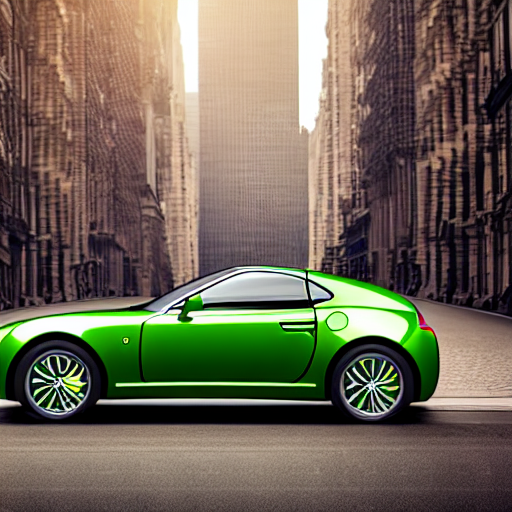}\\
\text{DB} \cite{Dreambooth} &
\text{TI} \cite{Textual} &
\text{TCO} \cite{TCO} &
\text{LoRA} \cite{LoRA} &
\text{CAT(\textbf{Ours})}\\
\end{tabular}
\begin{tabular}{c}
\includegraphics[width = 0.8in]{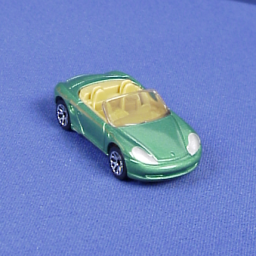} \\
\text{original} 
\end{tabular}

\begin{tabular}{ccccc}
\textit{with token} &
\textit{with token} &
\textit{with token} &
\textit{with token} &
\textit{with token}\\
\includegraphics[width = 1in]{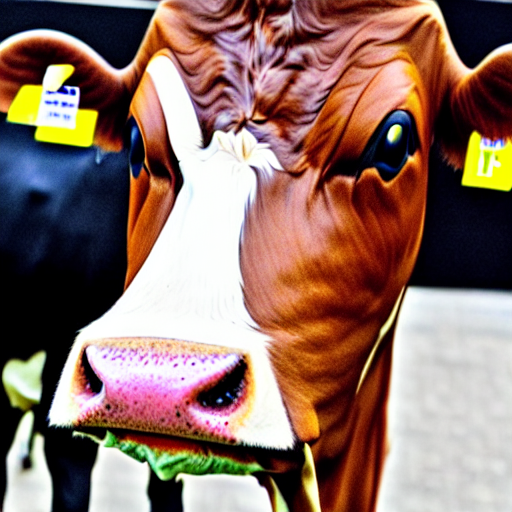} &
\includegraphics[width = 1in]{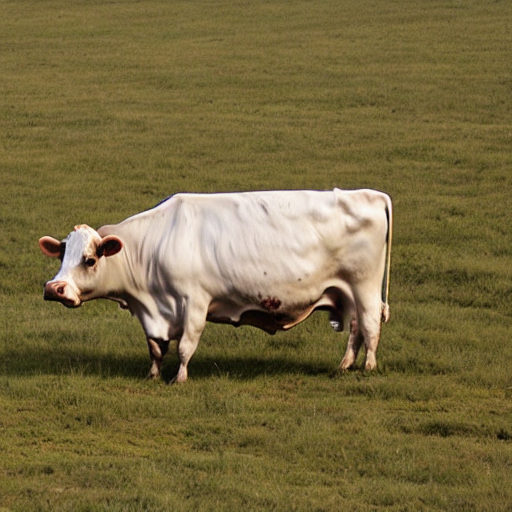} &
\includegraphics[width = 1in]{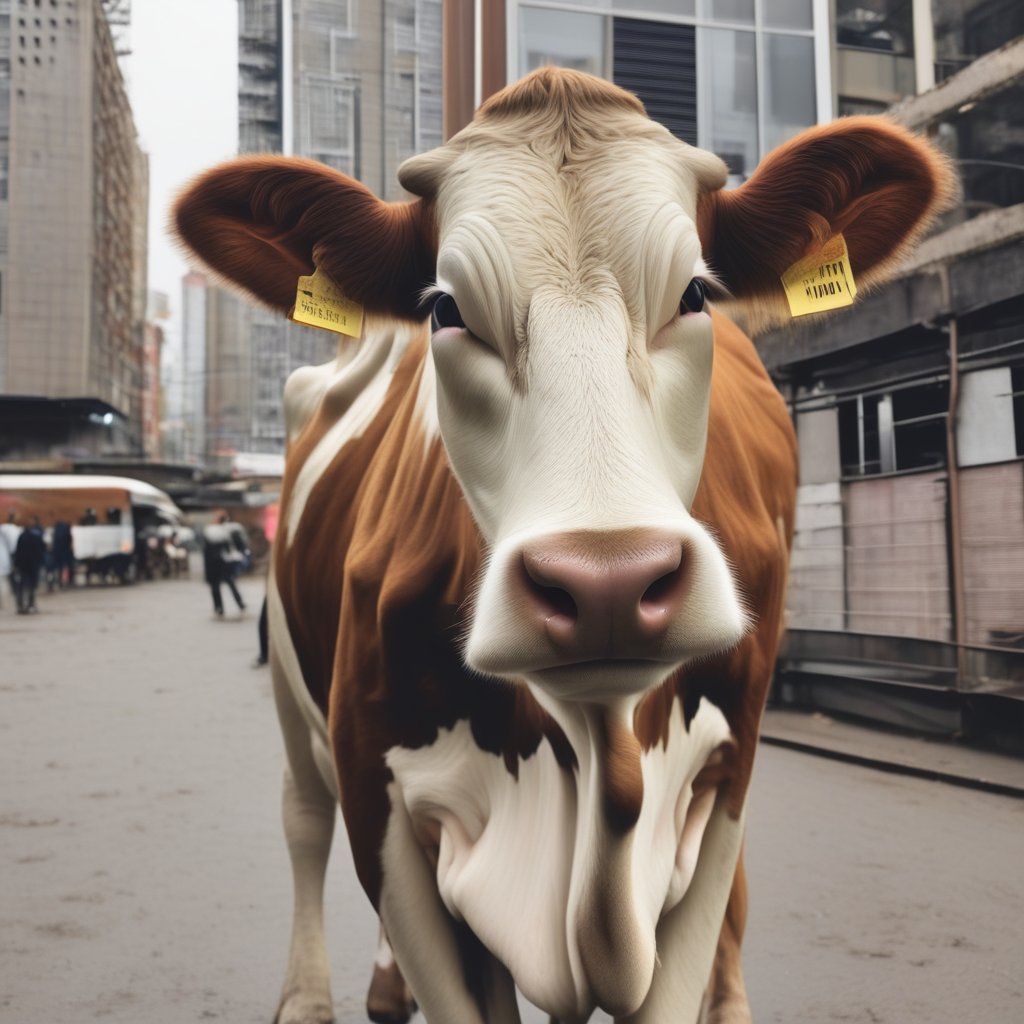} &
\includegraphics[width = 1in]{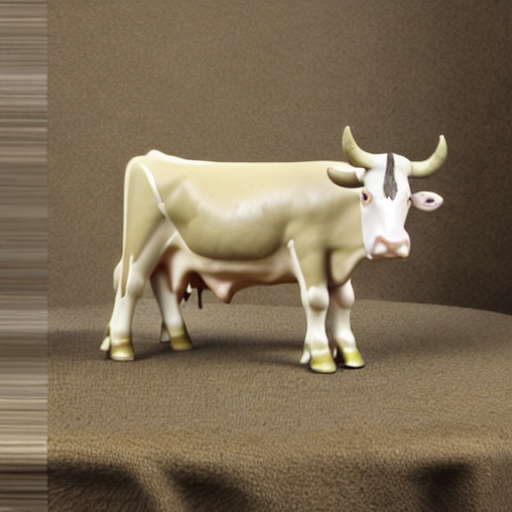} &
\includegraphics[width = 1in]{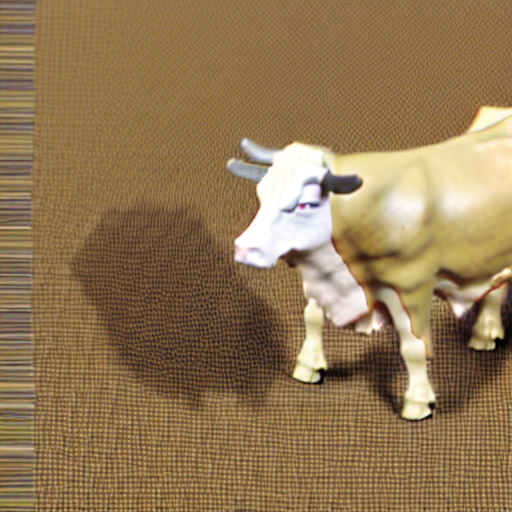}\\
\textit{no token} &
\textit{no token} &
\textit{no token} &
\textit{no token} &
\textit{no token}\\
\includegraphics[width = 1in]{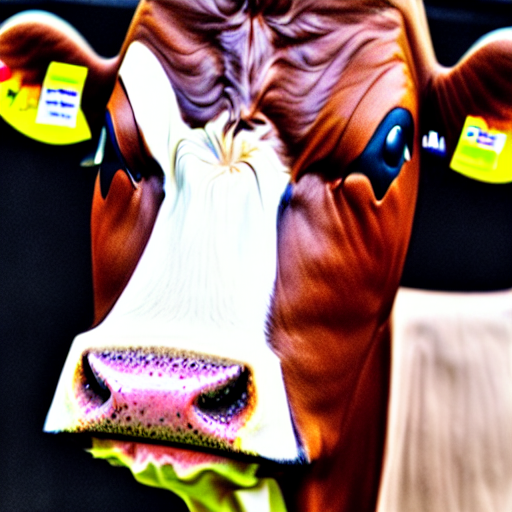} &
\includegraphics[width = 1in]{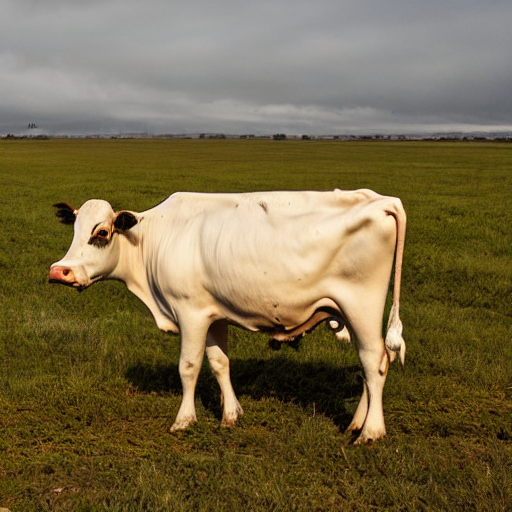} &
\includegraphics[width = 1in]{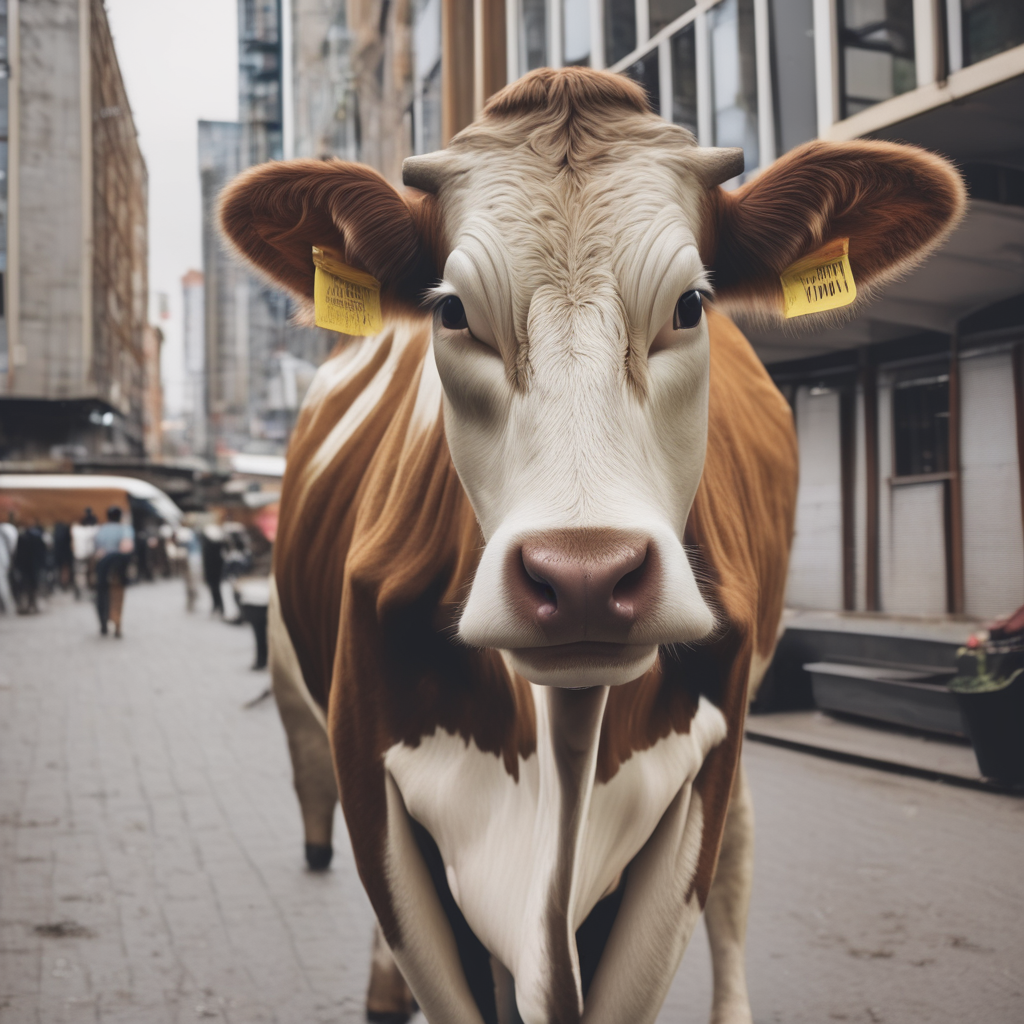} &
\includegraphics[width = 1in]{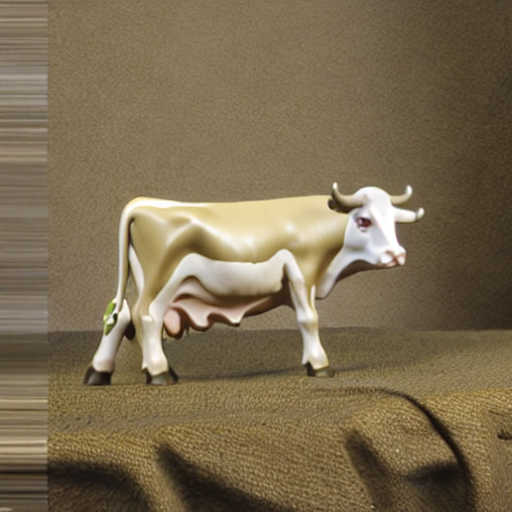} &
\includegraphics[width = 1in]{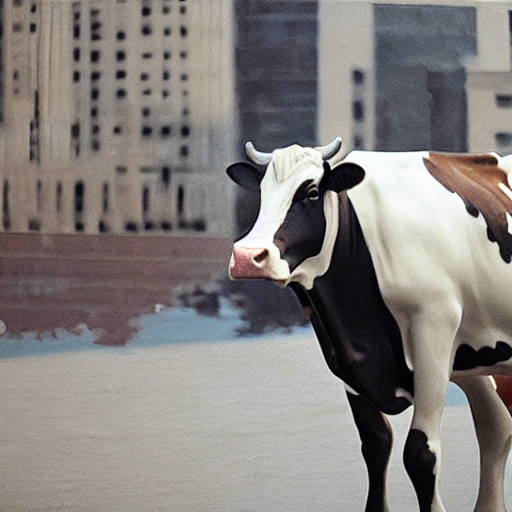}\\
\text{DB} \cite{Dreambooth} &
\text{TI} \cite{Textual} &
\text{TCO} \cite{TCO} &
\text{LoRA} \cite{LoRA} &
\text{CAT(\textbf{Ours})}\\
\end{tabular}
\begin{tabular}{c}
\includegraphics[width = 0.8in]{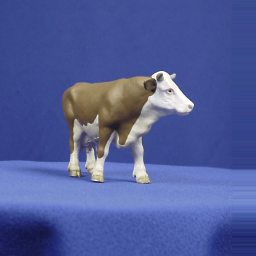} \\
\text{original} 
\end{tabular}

\begin{tabular}{ccccc}
\textit{with token} &
\textit{with token} &
\textit{with token} &
\textit{with token} &
\textit{with token}\\
\includegraphics[width = 1in]{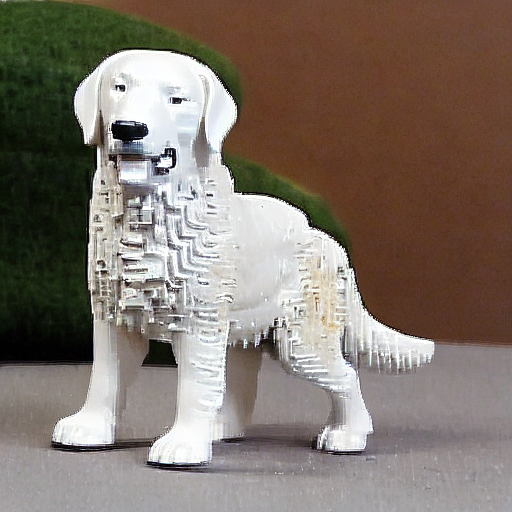} &
\includegraphics[width = 1in]{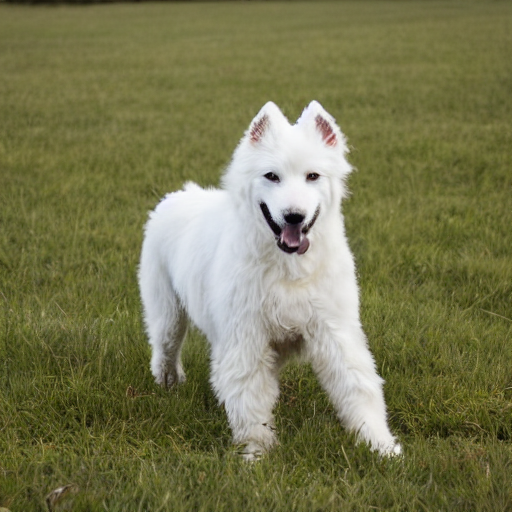} &
\includegraphics[width = 1in]{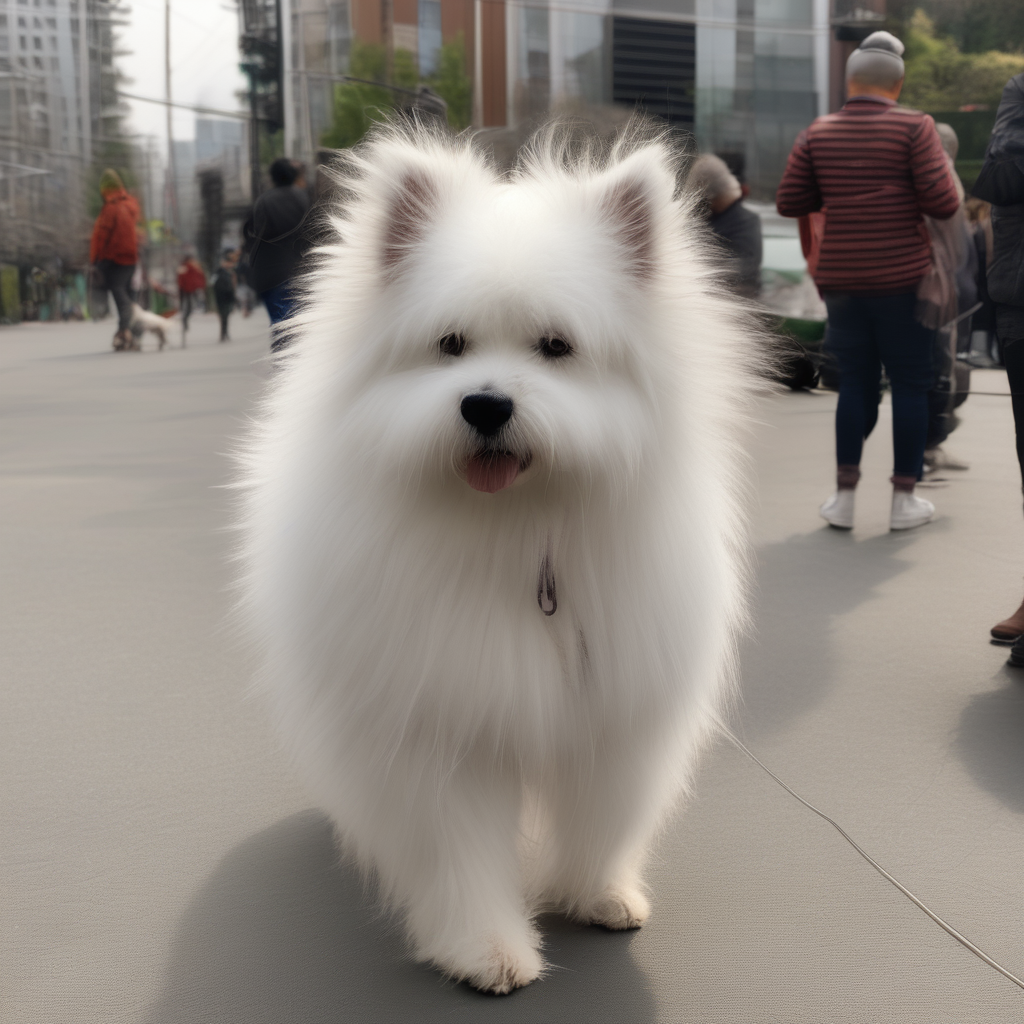} &
\includegraphics[width = 1in]{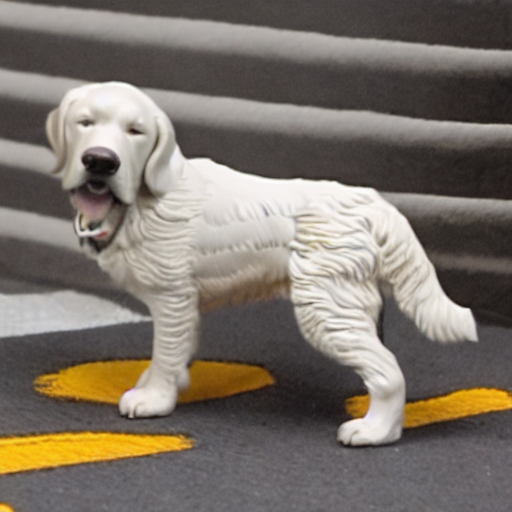} &
\includegraphics[width = 1in]{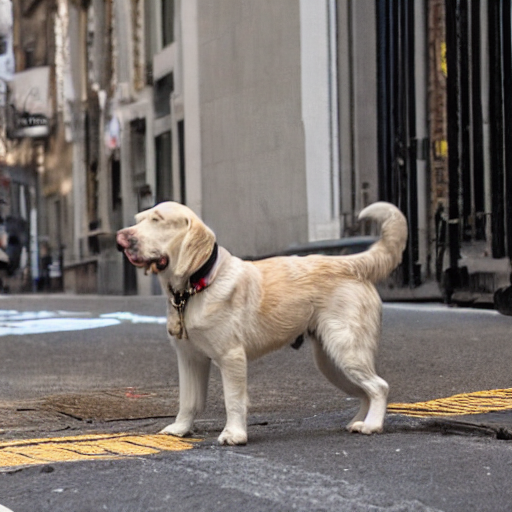}\\
\textit{no token} &
\textit{no token} &
\textit{no token} &
\textit{no token} &
\textit{no token}\\
\includegraphics[width = 1in]{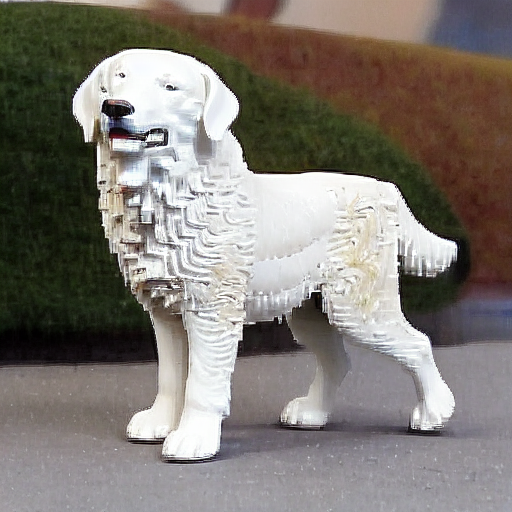} &
\includegraphics[width = 1in]{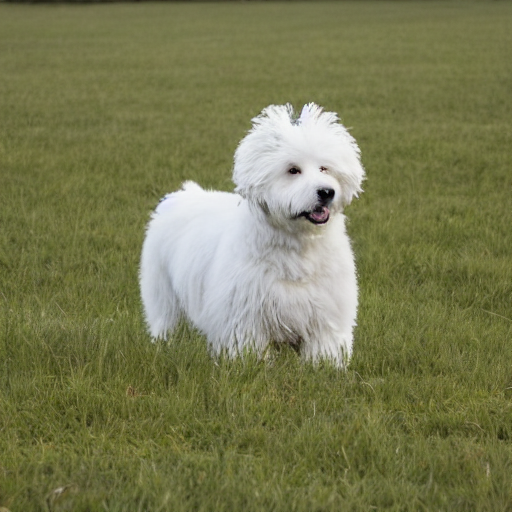} &
\includegraphics[width = 1in]{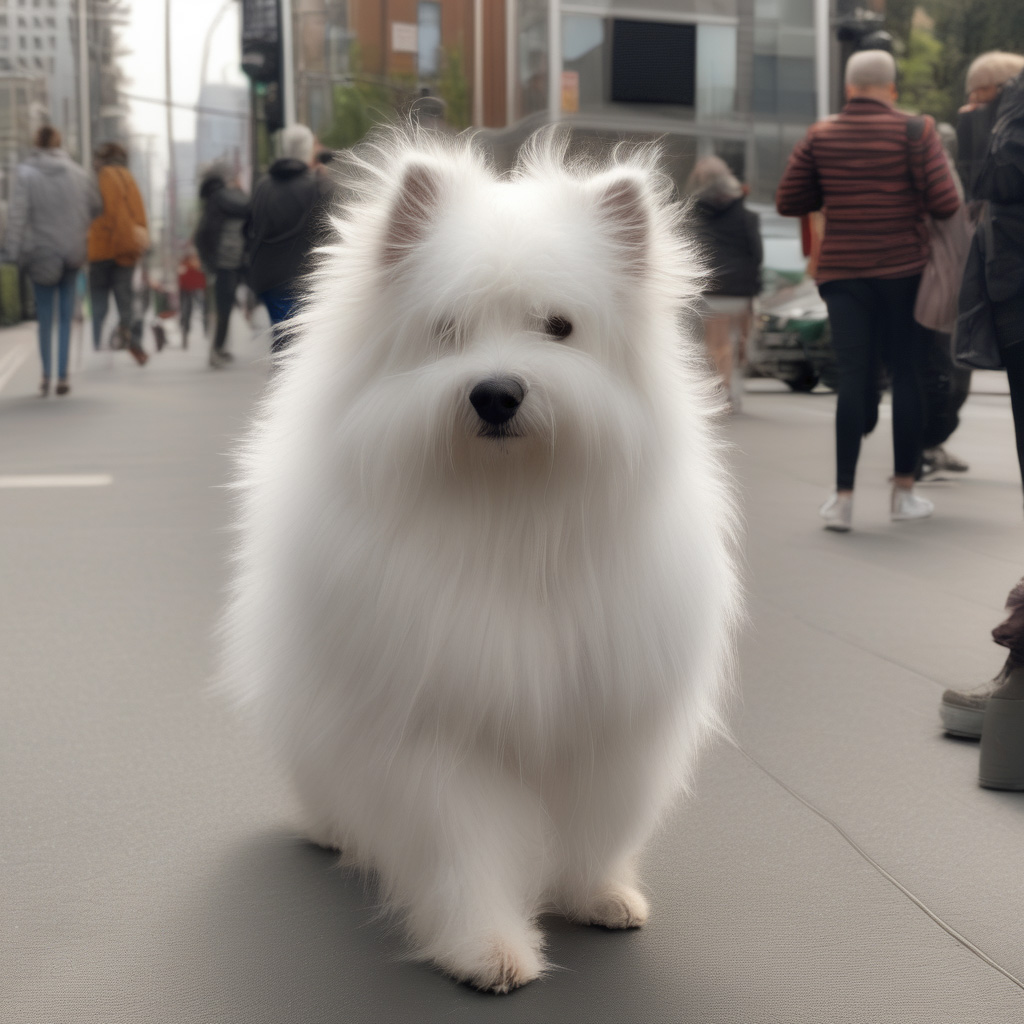} &
\includegraphics[width = 1in]{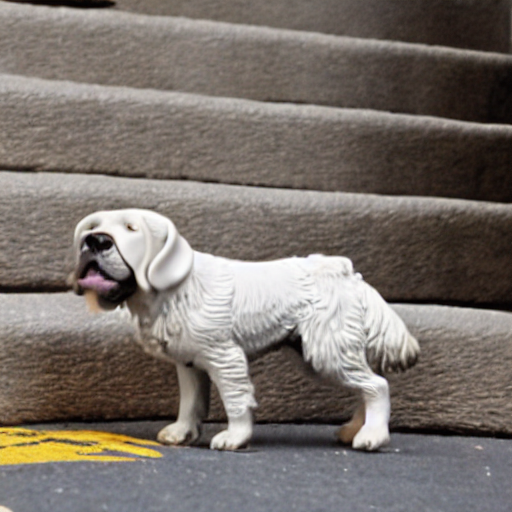} &
\includegraphics[width = 1in]{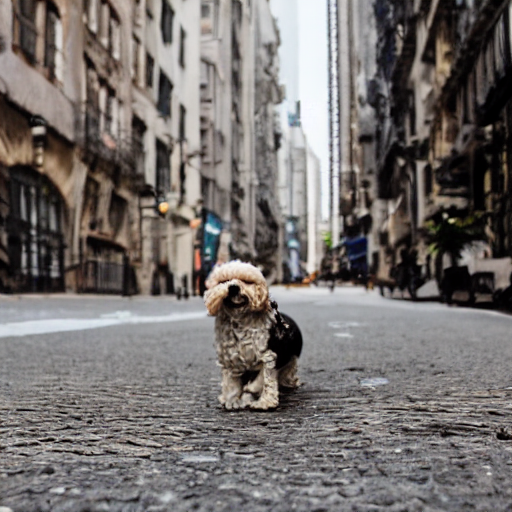}\\
\text{DB} \cite{Dreambooth} &
\text{TI} \cite{Textual} &
\text{TCO} \cite{TCO} &
\text{LoRA} \cite{LoRA} &
\text{CAT(\textbf{Ours})}\\
\end{tabular}
\begin{tabular}{c}
\includegraphics[width = 0.8in]{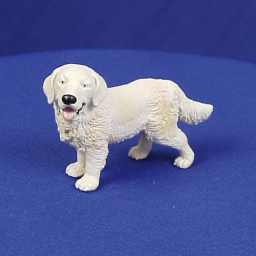} \\
\text{original} 
\end{tabular}

\captionof{figure}{\textbf{Qualitative results of various adapter training.} We used ETH-80 dataset\cite{eth} for training and validation. The prompts used for generation are \textit{a green sports car, a cow in a city, }and \textit{a dog in a city.} } 
\label{figure 4}
\end{figure*}

\section{Background}

\subsection{Mathematical Notations}
\label{math}
We denoted cosine similarity used in equation \ref{eq4}, \ref{eq6}, and \ref{eq8} as 

\begin{equation}
S_{C}(A,B) = \dfrac {A \cdot B} {\left\| A\right\| _{2}\left\| B\right\| _{2}} 
\end{equation}

For harmonic mean noted in \ref{prompt}, \ref{identity}, and \ref{kps},

\begin{equation}
H(x_1,x_2,\dots,x_n) = \dfrac{n}{\sum_{i=1}^{n} \dfrac{1}{x_i}}
\end{equation}

\subsection{Train Settings}
\label{train}
\begin{figure*}[h]
    \begin{subfigure}{0.5\textwidth}
        \centering
        \includegraphics[width=\linewidth]{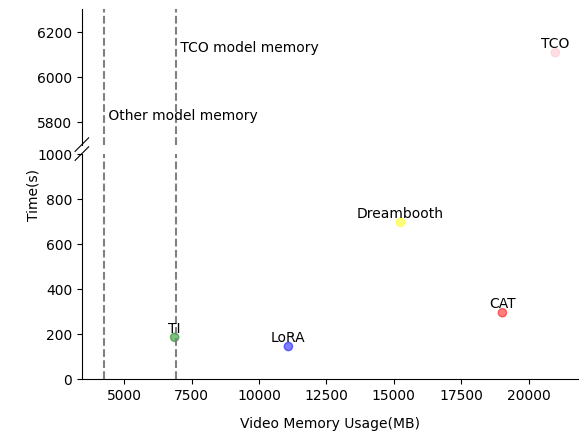}
        \caption{Resource usage in train}
    \end{subfigure}
    \begin{subfigure}{0.5\textwidth}
        \centering
        \includegraphics[width=\linewidth]{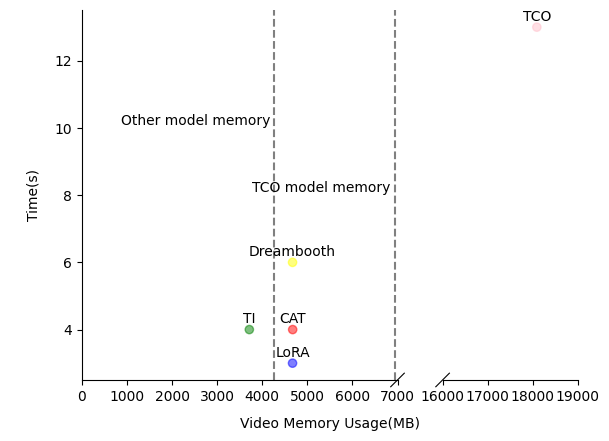}
        \caption{Resource usage in inference}
    \end{subfigure}
    \caption{\textbf{Video memory and time consumption of various adapters in train and inference.} The \textcolor{red}{red dots} show our method's performance. TCO \cite{TCO} uses Stable Diffusion XL model\cite{SDXL} requiring a larger VRAM consumption. We included the preparation time for regularization set with Dreambooth\cite{Dreambooth} and TCO in the total training time. We plan to conduct further optimizations to decrease memory consumption on the same level as LoRA \cite{LoRA}.}
    \label{fig:vram_graph}
\end{figure*}

For Figures \ref{fig: Figure 0}, \ref{figure 2}, \ref{fig:vram_graph} and Table \ref{tab:Table 1}, we utilized Textual Inversion dataset to validate our training. Each dataset was composed of 5-6 images. For the table, we trained total 9 adapters for each type and generated 10-11 images for each adapter to evaluate the metrics. For Figure \ref{figure 4}, we collected 6 images from each ETH-80 dataset for the experiment. For these inferences, we trained all adapters with equal 600 steps with mentioned datasets except the regularization sets which were an essential part of some adapter's algorithm. For Figure \ref{fig:Figure 4}, we only used a single dataset of \textit{red teapot} among the datasets above and trained 4 adapters with different CAT variables, $\alpha$, and 1000 steps. We applied uniform learning rate of 1e-4 for U-Net LoRA, Dreambooth and Textual Inversion. We applied AdamW optimizer \cite{ADAMW} with weight decay ${\lambda}$=0.01 and betas ${\beta}$=(0.9, 0.999), epsilon ${\epsilon}$=1e-08, utilizing max 18GB of VRAM in average, as described in Figure \ref{fig:vram_graph} with detailed training time. For hardware, we used 2 GeForce RTX 4090 gpus for all training.

\subsection{Multi-Concept Training}
\label{MultiConcept}
\begin{figure}[h]
    \centering
    \includegraphics[width=6in]{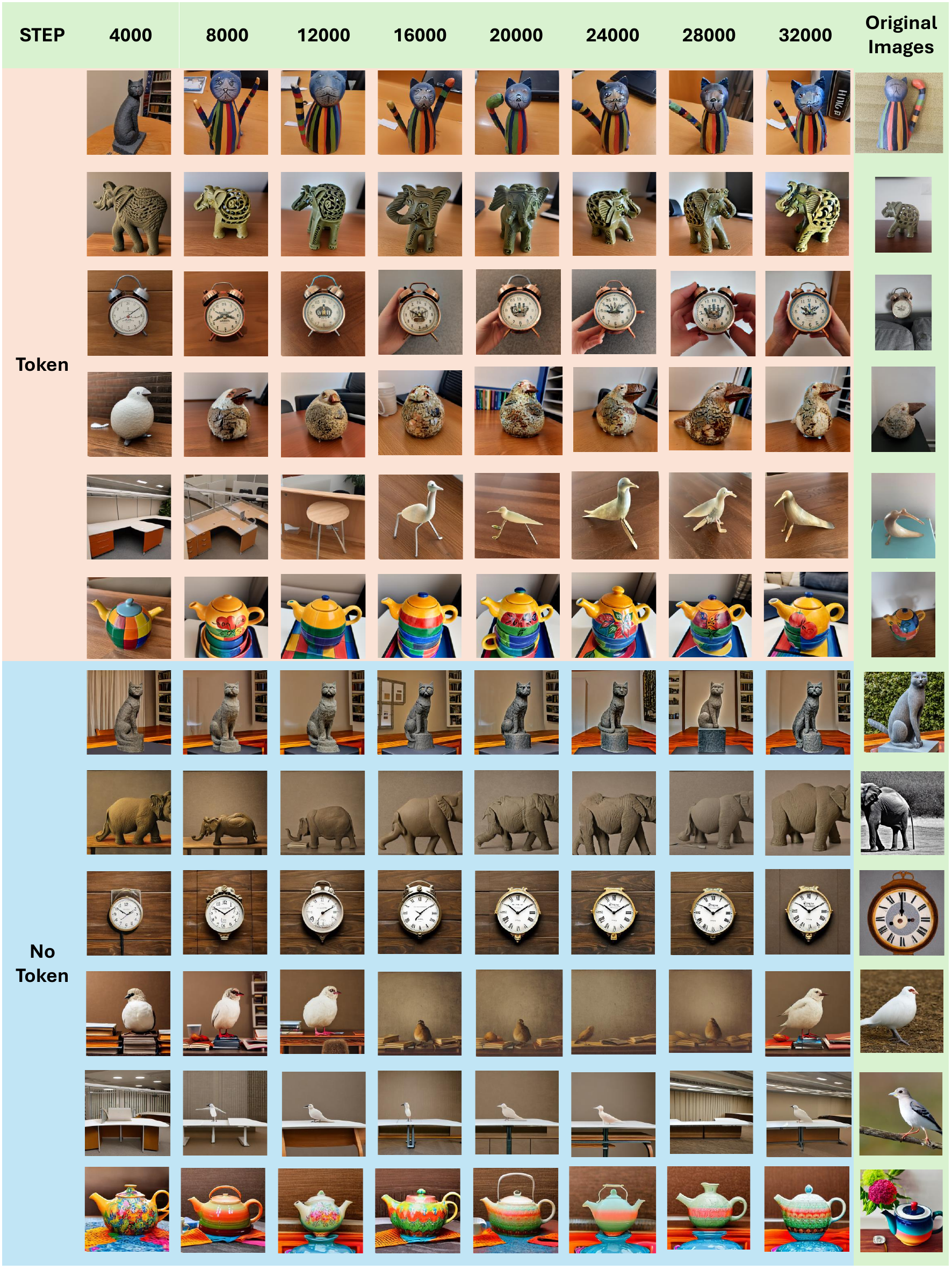}
    \caption{\textbf{Single Adapter Multi-Concept training samples.} The result shows the capability of learning multiple concepts as desired special tokens in single adapter.}
    \label{fig:multi_token}
\end{figure}

\begin{figure*}[h]
    \centering
    \includegraphics[width=\textwidth]{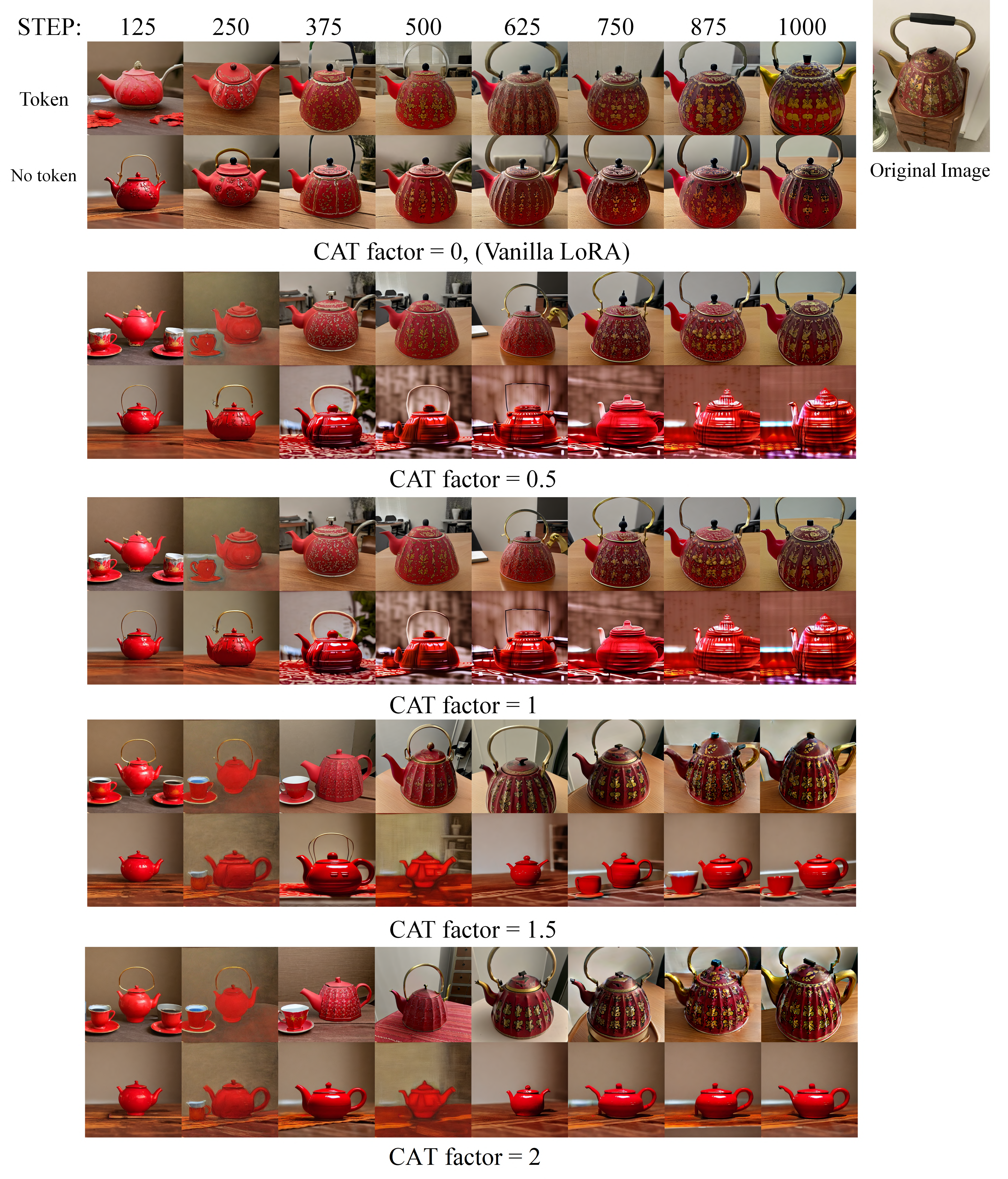}
    \caption{\textbf{The qualitative result of CAT factor experiment using LoRA\cite{LoRA}.} When the CAT factor is low, the original class generation tends to be off-grade, which is known as hyperspherical energy diverge \cite{OFT}. On the other hand, with proper tuning of CAT factor, the adapter generates the original class with better quality, also the injection of the original knowledge accelerates.}
    \label{fig:Figure 4}
\end{figure*}

For Figure \ref{fig:multi_token}, we selected 44 images from Textual Inversion to compose a single dataset. We trained only one LoRA based CAT adapter with total 32,000 steps and the same setting from Section \ref{train}. The experiment is to elucidate multi-concept training capability. This showed the result of a successful training trend of custom concepts in a single adapter, suggesting a large scale knowledge injection capability. For simplicity, we fixed all trigger tokens as first phrase, letting them be replaced dynamically.

\begin{figure}[h]
    \begin{subfigure}{0.5\textwidth}
        \centering
        \includegraphics[width=\linewidth]{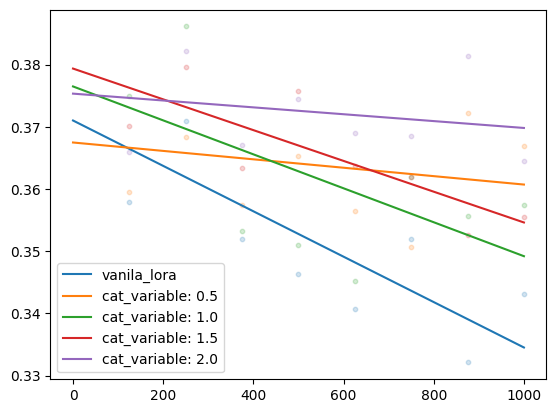}
        \caption{Prompt Score(\ref{prompt}) Graph regarding the experiment in Figure \ref{fig:Figure 4}}
    \end{subfigure}
    \begin{subfigure}{0.5\textwidth}
        \centering
        \includegraphics[width=\linewidth]{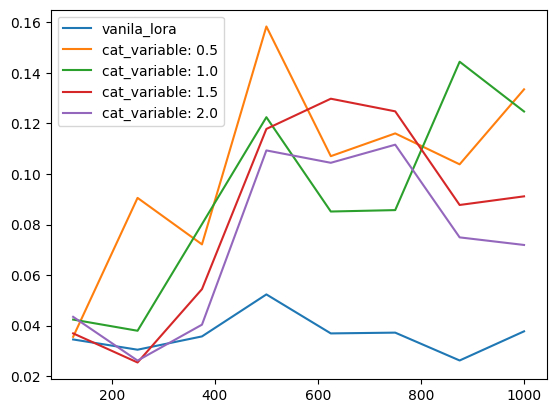}
        \caption{KPS(\ref{kps}) Graph regarding the experiment in Figure \ref{fig:Figure 4}}
    \end{subfigure}
    \caption{\textbf{Metric Graph on CAT variable test}. We used the metrics mentioned in Section \ref{eval} to evaluate the experiment in Figure \ref{fig:Figure 4}. As stated in the experiment, prompt score shows the higher capability of original knowledge generation. Also, KPS is always higher than the baseline LoRA as long as we apply CAT despite the variable's magnitude.}
    \label{fig:graph2}
\end{figure}

\end{document}